\documentclass[doc]{apa6}

\usepackage{listings}
\usepackage{float}

\usepackage{amsmath,amssymb}
\usepackage{mathtools}

\usepackage{changepage}

\usepackage[utf8x]{inputenc}

\usepackage{textcomp,marvosym}

\usepackage[numbers, super, sort]{natbib}

\usepackage[table]{xcolor}

\usepackage{array}

\usepackage{subcaption}

\title{A Method of Selective Attention for Reservoir Based Agents}
\shorttitle{Selective Attention for Reservoir Based Agents}
\author{Kevin McKee}
\affiliation{Astera Institute}

\abstract{
Training of deep reinforcement learning agents is slowed considerably by the presence of input dimensions that do not usefully condition the reward function.
Existing modules such as layer normalization can be trained with weight decay to act as a form of selective attention, i.e. an input mask, that shrinks the scale of unnecessary inputs, which in turn accelerates training of the policy.
However, we find a surprising result that adding numerous parameters to the computation of the input mask results in much faster training.
A simple, high dimensional masking module is compared with layer normalization and a model without any input suppression.
The high dimensional mask resulted in a four-fold speedup in training over the null hypothesis and a two-fold speedup in training over the layer normalization method.
}

\begin{document}
\maketitle
\section*{Introduction}
Natural intelligence relies on selective attention, the cognitive process of excluding unnecessary distractions from consideration and enhancing the most important information.
Several influential works in cognitive science have previously espoused the theory that attention arises from the brain's need to conserve resources\cite{broadbent2013perception, treisman1980feature, kahneman1973attention}.
However, selective attention is likely more than just a matter of resource constraints, and instead concerns the efficiency of learning by flexibly prioritizing information.\cite{dayan2000learning}
Four functions of selective attention in animals were identified through a review of empirical results: \cite{treisman1969strategies} 
\textit{``(1) restriction of the number of inputs analyzed, (2) restriction of the dimensions analyzed, (3) the items (defined by sets of critical features) for which [the subject] looks or listens, and (4) selections of which results of perceptual analysis will control behavior and be stored in memory.''}
The implication of this theory is that there may exist general methods of intelligently suppressing uninformative inputs to improve the efficiency of machine learning.
In the reinforcement learning (RL) context, reward signals present a sufficiently general objective for such a filter.

In this study, we consider the model of selective attention as suppression of unimportant inputs to improve the efficiency of learning.
We test for changes in reinforcement learning efficiency as a result of both existing method and novel methods.
Our results demonstrate, first, that input suppression controlled by reward signals greatly accelerates learning, and second, that greatly over-parameterizing the input suppression mask results in the largest speedups in training.
We give this new method the nickname Excessively Parameterized Input Concealment, or EPIC.

Specifically, we develop this method for use with reinforcement learning agents that incorporate reservoir computing. 
In previous work, it was found that a reservoir computer formulated as a recurrent neural network (RNN) with fixed, random, and normalized weights, called the Echo State Network (ESN), outperforms other memory architectures on experimental tasks that require meta-learning and non-Markovian time dependence.\cite{mckee2024reservoir}
However, reservoir computers do not typically allow for the training of input weights to the RNN.
Training a multi-layer decoder with back-propagation therefore scales poorly as number of inputs dimensions with non-trivial variances increases.
By incorporating reward-driven selective attention, inputs can be either suppressed or amplified before they are registered in the RNN, resulting in more efficiently decoded memory representations with a larger signal-to-noise ratio with respect to the final rewards.

\section*{Task}
\paragraph{Distracting Multi-armed Bandit}
The primary task involved a standard Multi-armed bandit as detailed in both ours and others' previous experiments with meta-learning,\cite{mckee2024reservoir, wang2018prefrontal} with the addition of a large observation vector of random, normally distributed noise.
Half of the observation vector was updated with new noise every step, while the other half updated only once per episode.
The noise vector served to artificially complicate the loss function and imitate the case of high-dimensional real-world distractions to the agent.

\section{Models}
All agents used the actor critic algorithm regularized with entropy maximization.
The design was the same as used in previous related work. \cite{mckee2024reservoir}
In short, inputs to the agent and action-reward feedbacks are passed to the Echo State Network (ESN), which then feeds forward through two-layer MLPs for each the actor and the critic outputs.
The Echo State Network incorporated a dense-local and sparse-global connection pattern.
The MLPs had hidden layers with 256 nodes each.
Hyperparameters are given in Appendix A.

\paragraph{Layer Normalization}
Layer normalization with element-wise affine transformation (LayerNorm)\cite{ba2016layer} provides a common sense solution for suppressing inputs.
\begin{align}
    y &= \frac{\mathbb{E}[X]}{\sqrt{\text{Var}(X)+\epsilon} }\odot \gamma + \beta
\end{align}
To test the effectiveness of layer normalization as an input suppressant, one agent included a standard layer normalization step between the inputs and the ESN. 
The initial scale values were multiplied by 2.5, to allow a broader range of inputs to the ESN.
This minor step was found through manual search to improve training time.
To use layer normalization to suppress inputs, a relatively large weight decay coefficient (1e-4) was included in the loss function only with respect to the parameters $\gamma$ and $\beta$.

\paragraph{Simple vector filter}
A simple alternative to layer normalization is to use a simple real valued vector of parameters $b$.
The parameters are transformed into a mask by bounding in the $[\min, \max]$ interval as follows:
\begin{align}
    m &= (\max-\min)\sigma (b) + \min, \\
    y &= x \odot m.
\end{align}
This represents the simplest filtering strategy tested as it uses the fewest parameters and unlike layer normalization, involves no normalization step.

\paragraph{Excessively Parameterized Input Concealment}
A less common sense solution to suppressing inputs is to start with a fixed vector of random numbers $u$ several times length of the inputs $x$, then applying a trained affine transformation of those random numbers to produce the mask values $m$.
The mask values are bound using a scaled and shifted sigmoid function:
\begin{align}
    u &\sim N(0,1),\\
    m &= (\max-\min)\sigma (Wu + b) + \min, \\
    y &= x \odot m,
\end{align}
where $W$ is a fully connected weight matrix and $b$ is the bias term.
This strategy arbitrarily over-parameterizes the mask so that during training, there are many equivalent solutions, and many that will, by chance, be close to the parameter's starting vector.
By making the mask's training as speedy as possible, the search space for the actor and critic parameters is greatly reduced for the remaining majority of training.

This strategy requires the actor and critic modules backpropagate error gradients to the mask generating parameters.
Those gradients must be larger than the gradients of a regularizing penalty that shrinks the mask.
That penalty is nothing more than the mean value of the final mask, weighted by a small scaler.
Among values manually tried for these experimental tasks, the best value for that scaler was found to be 1e-5.
For larger input tasks, it may be necessary to tune this coefficient.

\section{Results}
\begin{figure}[ht]
    \centering
    \begin{subfigure}[t]{0.48\textwidth}
        \includegraphics[width=\linewidth]{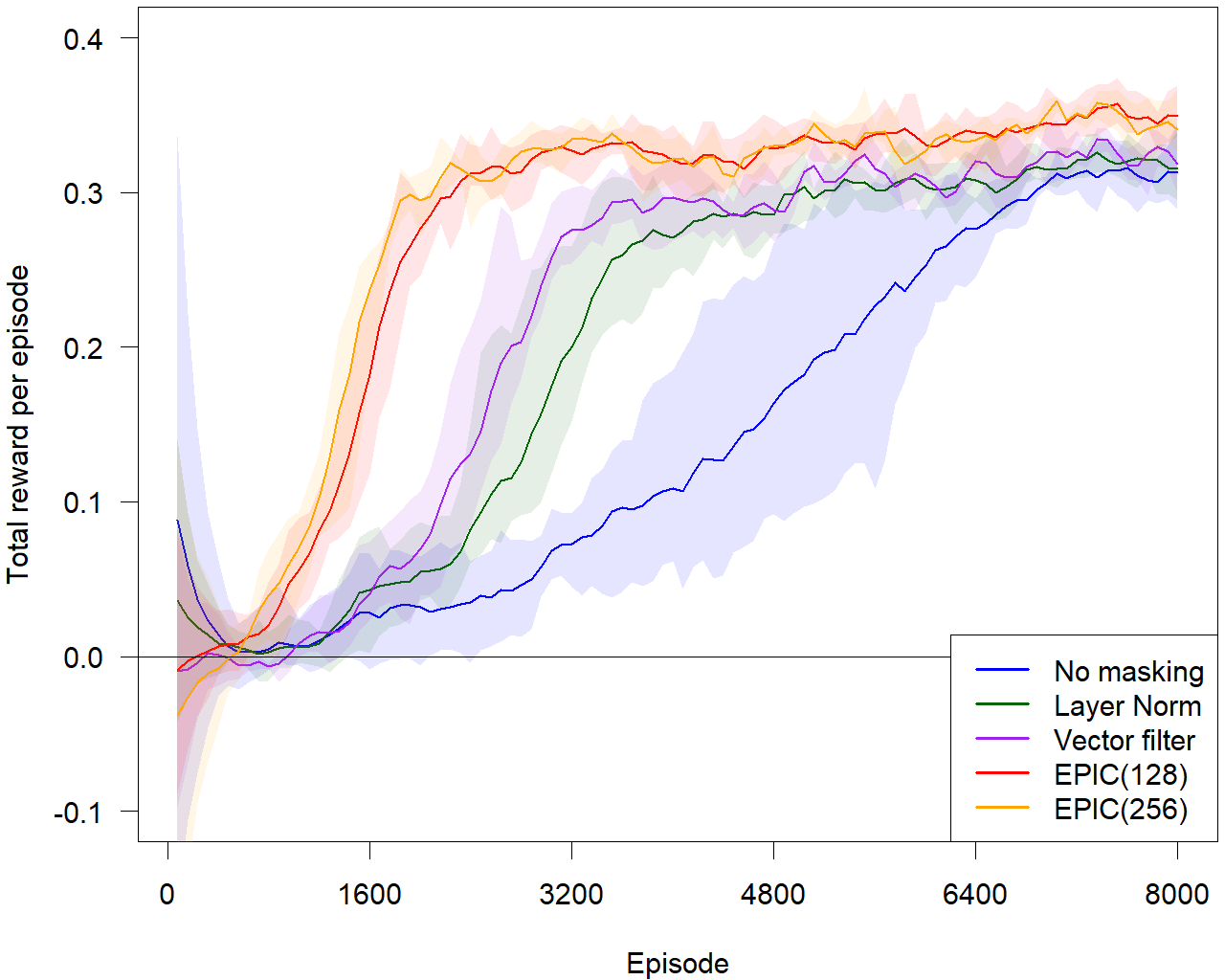}
        \caption{Results for multi-armed bandit with 32-length distraction noise.}
        \label{fig:results_mab_32}
    \end{subfigure}
    \begin{subfigure}[t]{0.48\textwidth}
        \includegraphics[width=\linewidth]{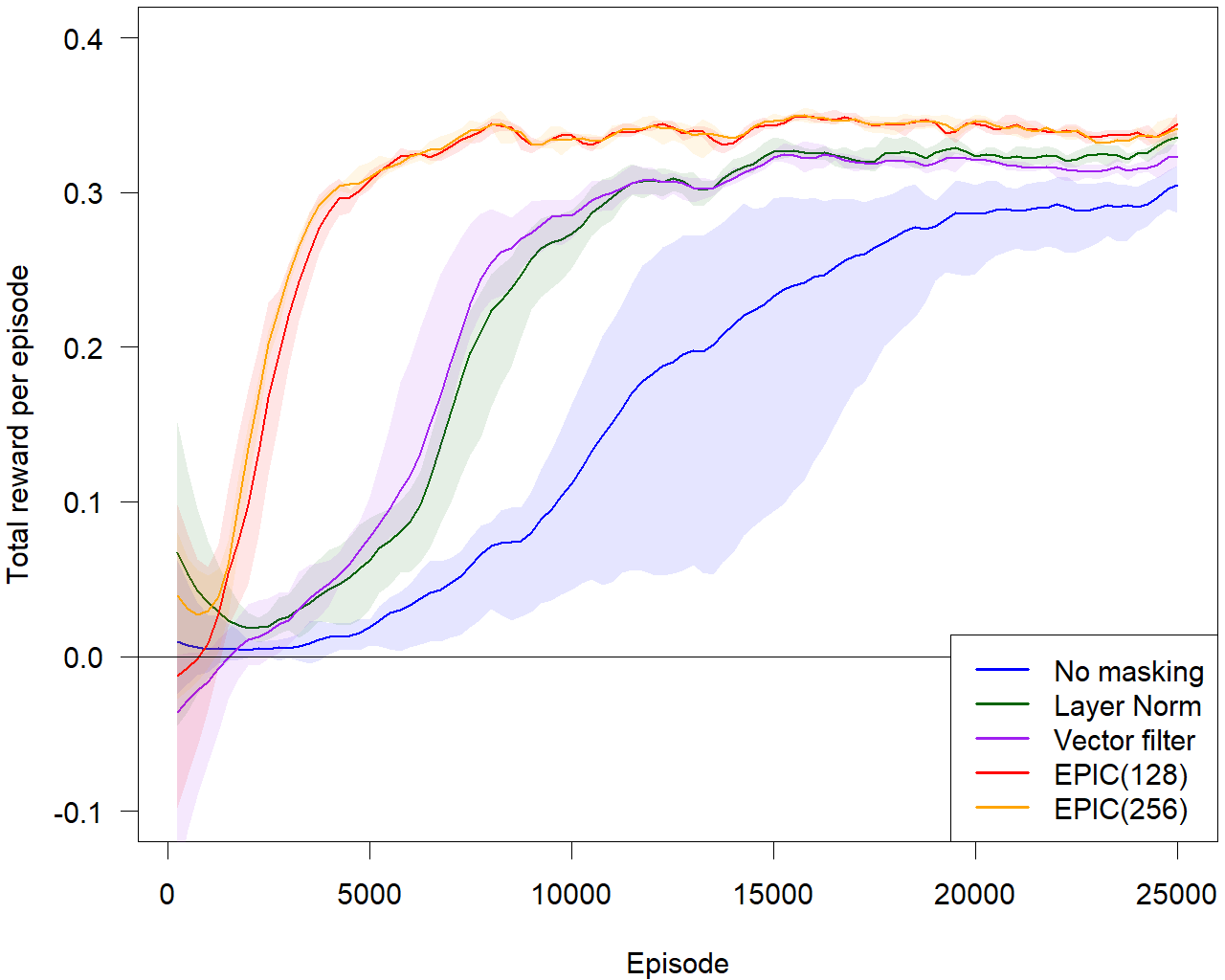}
        \caption{Results for multi-armed bandit with 64-length distraction noise.}
        \label{fig:results_mab_64}
    \end{subfigure}
    \caption{Results on multi-armed bandit with 32 and 64-length distraction noise.  Trained input masking by regularized layer normalization methods doubled training efficiency. Over-parameterizing the mask quadrupled training efficiency. For EPIC, the length of random vector $u$ is given in parentheses.}
\end{figure}

The results of this experiment are shown in Figure \ref{fig:results_mab_32}.
The distraction vector resulted in the ESN based agent reaching a high score in about 7000-8000 episodes, rather than the expected 500-1500 episodes. \cite{mckee2024reservoir}
Adding either the vector filter or layer normalization with weight decay halved the number of episodes to convergence.
Replacing either with EPIC halved the number of episodes again.
Increasing the size of the random vector in EPIC only made a negligible improvement in training time.
When the task had 64 (Figure \ref{fig:results_mab_64}) distraction inputs instead of 32 (Figure \ref{fig:results_mab_32}), the ratios of performance were about the same with a slightly more pronounced improvement in EPIC in the former.

Besides training time itself, the reliability of the training curves also appears to improve with better input masking.  
The model without any input mask varied in its convergence, ranging from nearly the same rate as that with layer normalization, to taking twice as many episodes. 

\section{Discussion}
Our results demonstrate first that intelligently masking the inputs results in much faster learning in the presence of many extra, uninformative input dimensions.
This was done in both of the examined methods by balancing reward gradients with an objective to minimize the scale of the inputs.
Second, we demonstrate a surprising principle that the seemingly unnecessary over-parameterization of the input mask causes significant further improvements to training efficiency.
We hypothesize that this effect results from providing the input mask-generating function with many equivalent but randomly distributed solutions such that one is likely to exist close to the network's starting parameters at the beginning of training.
By making such solutions convenient for the solver, the mask can reduce the search space for the policy and critic networks early in training, leading to the overall speedups observed.

\paragraph{Limitations}
Our hypothesis and findings arose in the context of developing an efficient reservoir-based RL agent, so our study was limited to that context. 
We do not examine here whether the input masking method is superior to alternatives in general, or superior to trainable input weights in general.
We only justify our result in light of the principles by which reservoir-based agents work.

In this paper, we do not examine the more general goal of conditional input masking, which better reflects naturalistic usage of selective attention.
While we leave conditional masking for future work, the current findings regarding over-parameterization were in fact a consequence of repeatedly finding that conditioning the mask on the recurrent (ESN) state via multiple, fully connected layers produced the improvements over fixed masks shown here.
When conditioning on the ESN state, the resulting mask nonetheless tended to demonstrate a fixed set of scalars over the input vector, leading to the current hypothesis that the excessive parameter dimensionality was more the cause of improvement than was conditioning on the ESN.
Furthermore, the task used here, and perhaps most familiar diagnostic tasks, tend to entail a fixed degree of importance per input element.
To study conditional selection of inputs, more work is required to conceptualize a set of tasks that require or emphasize the necessity to shift the mask's values with changing context.

\paragraph{Conclusion}
Combining reward signal backpropagation with regularization can result in learned, intelligent masking of inputs reflecting their importance to the task.
This masking reduces the search space for the policy and critic networks, resulting in significant improvements to training efficiency in agents that use reservoir based memory.
Over-parameterizing the input mask-generating function further increases the training efficiency by as much as double.
Hence the latter method represents an easy, generic, and low-cost strategy for improving RL performance.

\bibliographystyle{unsrtnat}
\bibliography{refs}

\begin{thebibliography}{8}
\providecommand{\natexlab}[1]{#1}
\providecommand{\url}[1]{\texttt{#1}}
\expandafter\ifx\csname urlstyle\endcsname\relax
  \providecommand{\doi}[1]{doi: #1}\else
  \providecommand{\doi}{doi: \begingroup \urlstyle{rm}\Url}\fi

\bibitem[Broadbent(2013)]{broadbent2013perception}
Donald~Eric Broadbent.
\newblock \emph{Perception and communication}.
\newblock Elsevier, 2013.

\bibitem[Treisman and Gelade(1980)]{treisman1980feature}
Anne~M Treisman and Garry Gelade.
\newblock A feature-integration theory of attention.
\newblock \emph{Cognitive psychology}, 12\penalty0 (1):\penalty0 97--136, 1980.

\bibitem[Kahneman(1973)]{kahneman1973attention}
Daniel Kahneman.
\newblock \emph{Attention and effort}, volume 1063.
\newblock Citeseer, 1973.

\bibitem[Dayan et~al.(2000)Dayan, Kakade, and Montague]{dayan2000learning}
Peter Dayan, Sham Kakade, and P~Read Montague.
\newblock Learning and selective attention.
\newblock \emph{Nature neuroscience}, 3\penalty0 (11):\penalty0 1218--1223, 2000.

\bibitem[Treisman(1969)]{treisman1969strategies}
Anne~M Treisman.
\newblock Strategies and models of selective attention.
\newblock \emph{Psychological review}, 76\penalty0 (3):\penalty0 282, 1969.

\bibitem[McKee(2024)]{mckee2024reservoir}
Kevin McKee.
\newblock Reservoir computing for fast, simplified reinforcement learning on memory tasks.
\newblock \emph{arXiv preprint arXiv:2412.13093}, 2024.

\bibitem[Wang et~al.(2018)Wang, Kurth-Nelson, Kumaran, Tirumala, Soyer, Leibo, Hassabis, and Botvinick]{wang2018prefrontal}
Jane~X Wang, Zeb Kurth-Nelson, Dharshan Kumaran, Dhruva Tirumala, Hubert Soyer, Joel~Z Leibo, Demis Hassabis, and Matthew Botvinick.
\newblock Prefrontal cortex as a meta-reinforcement learning system.
\newblock \emph{Nature neuroscience}, 21\penalty0 (6):\penalty0 860--868, 2018.

\bibitem[Ba et~al.(2016)Ba, Kiros, and Hinton]{ba2016layer}
Jimmy~Lei Ba, Jamie~Ryan Kiros, and Geoffrey~E Hinton.
\newblock Layer normalization.
\newblock \emph{arXiv preprint arXiv:1607.06450}, 2016.

\end{thebibliography}

\newpage 

\section{Appendix A: Hyperparameters}
The hyperparameters for the ESNs are given in Table \ref{tab:hyperpars}.

\begin{table}[!h]
    \centering
    \caption{Hyperparameters for all models, including fixed ESN weight matrices.}
    \begin{tabular}{l|l|l}
        \hline
        Parameter & Description & Value\\ 
        \hline
        \textbf{Actor-Critic} &  &\\
         $LR$ & Learning rate  & 0.0001\\
         $\beta_e$ & Entropy regularization coefficient & 0.001\\
         $N_\text{Hidden}$ & Hidden units per MLP layer & 256 \\
         $K$ & Number of hidden layers per MLP & 2 \\
         \hline
          \textbf{ESNLG} & &\\
         $N_{\text Unique}$ & Unique hidden nodes per input & 40\\
         $N_{\text Shared}$ & Overlapping nodes per neighboring inputs & 20\\ 
         $\phi$ & Spectral radius & 1.0 \\
         $P_L(W)$ & Local connection probability & 50\% \\
         $P_G(W)$ & Global connection probability & 1\% \\
         $P_I(W)$ & Input connection probability & 50\% \\
         $R$ & Max local connection radius & 10 \\
         \hline
          \textbf{EPIC} & &\\
         $N_u$ & Length of random normal vector & Input dimension $\times 4, 8$\\
          & Regularizer coefficient & $1e-5$\\ 
         min & Minimum mask value & 0.25 \\
         max & Maximum mask value & 5.0 \\
         \hline
          \textbf{Vector filter} & &\\
          & Regularizer coefficient & $1e-5$\\ 
         min & Minimum mask value & 0.25 \\
         max & Maximum mask value & 5.0 \\
         \hline
    \end{tabular}
    \label{tab:hyperpars}
\end{table}

\end{document}